\documentclass[letterpaper, 10 pt, conference]{ieeeconf}  % Comment this line out if you need a4paper

\IEEEoverridecommandlockouts                              % This command is only needed if 
\usepackage{graphicx}
\usepackage{multirow}
\usepackage[table]{xcolor}
\usepackage{array}
\usepackage{hhline}
\usepackage{makecell} 
\usepackage{subcaption}
\usepackage{booktabs}
\usepackage{pifont}
\usepackage{amsmath}

\usepackage{url}
\providecommand{\IEEEauthorblockN}[1]{#1}
\providecommand{\IEEEauthorblockA}[1]{#1}

\usepackage{tikz}

\newcommand{\fullcircle}{\tikz\fill (0,0) circle (3.1pt);}
\newcommand{\emptycircle}{\tikz\draw (0,0) circle (2.8pt);}
\newcommand{\halfcircle}{%
  \begin{tikzpicture}
    \path[fill] (0,0) -- (90:2.8pt) arc (90:270:2.8pt) -- cycle; % fill left half
    \draw (0,0) circle (3pt); % outline
  \end{tikzpicture}%
}
\newcommand{\redcircle}{\tikz\fill[black] (0,0) circle (3.1pt);}

\title{\LARGE \bf
\texttt{\LARGE{CADET}}: A Modular Platform for Evaluating Distributed Cooperative Autonomy in Connected Autonomous Vehicles
}

\author{
    \IEEEauthorblockN{Pragya Sharma, Brian Wang, Mani Srivastava$^\dag$ \thanks{$\dag $Author holds concurrent appointments as a Professor of ECE and CS (joint) at UCLA, and as an Amazon Scholar. This paper describes work performed at the UCLA, and is not associated with Amazon.}} \\
    \IEEEauthorblockA{\textit{University of California Los Angeles}} \\
    \IEEEauthorblockA{\textit{\{pragyasharma, wangbri1, mbs\}@ucla.edu}}
}

\begin{document}

\maketitle
\thispagestyle{empty}
\pagestyle{empty}

%%%%%%%%%%%%%%%%%%%%%%%%%%%%%%%%%%%%%%%%%%%%%%%%%%%%%%%%%%%%%%%%%%%%%%%%%%%%%%%%
% \begin{abstract}

% Autonomous vehicles (AV) increasingly rely on deep learning pipelines for real-time perception and decision-making. However, deploying these pipelines across heterogeneous compute platforms--ranging from onboard systems to edge servers and cloud datacenters--introduces non-trivial tradeoffs among inference latency, accuracy, energy consumption, and overall downstream task performance. While prior research has evaluated these pipelines in isolation, there remains a critical lack of tooling for conducting end-to-end, distributed inference experiments in realistic AV scenarios.

% We present CADET, the CARLA-based Distributed Experimentation Toolkit, a modular framework built on top of CARLA and Scenario Runner for systematic, reproducible evaluation of distributed AV pipelines. CADET integrates a client-server abstraction to simulate perception offloading across device, edge, and cloud configurations, and supports automated measurement of model-level, system-level, and performance-level metrics. CADET further incorporates NetWaggle, a Mininet-based emulation subsystem for injecting realistic network conditions, including delay, jitter, and congestion, to simulate runtime variability in data offloading.

% % CADET supports a broad range of usage tiers, from robotics researchers studying closed-loop control under delay, to systems researchers benchmarking inference workloads, to ML practitioners automating scenario generation and dataset collection.

% \end{abstract}

\begin{abstract}
Deep learning models are increasingly central to autonomous vehicle (AV) pipelines, yet their integration has traditionally followed a monolithic design where perception, planning, and control execute on a single onboard computer. This design overlooks the emerging paradigm of cooperative autonomy, where vehicles interact with roadside units (RSUs), edge servers, and cloud-hosted intelligence through vehicle-to-everything (V2X) connectivity. Cooperative perception and control improve safety and efficiency, but also introduce systems-level challenges: network latency, compute heterogeneity, and multi-tenant contention, all critically affect real-time decision-making. These challenges are further amplified by the increasing reliance on large foundation models, whose scale necessitates cloud deployment.

We present CADET (Cooperative Autonomy through Distributed Experimentation Toolkit), a modular platform for systematic and reproducible evaluation of distributed cooperative autonomy systems under realistic deployment conditions. CADET decouples the AV stack into composable modules that can be flexibly deployed across vehicles, infrastructure, and edge/cloud tiers. The framework integrates state-of-the-art models, incorporates trace-driven network and workload emulation, and provides synchronized model-, system-, and task-level instrumentation. Through V2V and V2I experiments, we show that distributed deployment choices fundamentally shape safety, with V2V intent packets outperforming cloud-based perception and RSU-assisted perception sustaining safety until overloaded by concurrent requests. Although designed for AV pipelines, CADET also supports dataset-driven experimentation, enabling systems and ML researchers to benchmark distributed inference workloads independently of full vehicle simulation. CADET is open source, with code and demo available at https://nesl.github.io/cadet-web.

% By bridging autonomy algorithms with deployment realities, CADET enables rigorous evaluation of cooperative autonomy and offers a general-purpose testbed for distributed ML systems.
\end{abstract}

\section{Introduction}

Autonomous vehicles (AV) increasingly rely on deep learning models for perception, planning, and control. As these models have grown in scale and capability, they have been integrated into platforms such as Apollo~\cite{apollo2017}, Autoware~\cite{autoware}, and Pylot~\cite{gog2021pylot} to create comprehensive AI-driven AV pipelines. While modular in design, enabling researchers to swap components and experiment with novel algorithms, these platforms remain monolithic in deployment. They execute entire pipelines on single onboard computers, constraining experimentation to fixed hardware configurations and isolated vehicle intelligence.

Recent advances in vehicle-to-everything (V2X) connectivity are reshaping the assumption of siloed AV stack execution. The emerging paradigm of cooperative autonomy (CA) allows vehicles to exchange sensor data, predictions, and control advisories with other vehicles, roadside units (RSUs), and edge/cloud infrastructure~\cite{caillot2022survey, cp-nw-cui2022coopernaut}. Such cooperation has been shown to improve safety under occlusion~\cite{safety-aoki2019v2v, safety-soto2022survey}, improve traffic efficiency~\cite{milanes2014modeling}, and reduce collision probability in dense environments. However, realizing these benefits depends not merely on the algorithmic accuracy of individual deep learning models but on complex systems-level interactions. The same cooperative perception algorithm can yield different safety outcomes depending on computational resources and communication reliability.

The performance of cooperative autonomy is shaped by three key design-time choices: (i) model architecture selection which determines which models to use in the AV pipeline (ranging from lightweight CNN detectors~\cite{yolo11_ultralytics} to large transformer-based predictors~\cite{transformer}); (ii) deployment topology which specifies where these models are executed (on-vehicle, at RSUs, or on edge/cloud servers, each with distinct latency--accuracy trade-offs~\cite{edge-cloud-li2018jalad}); and (iii) network characteristics which govern how these components communicate (wireless jitter, congestion, and multi-client interference that introduce non-deterministic delays and directly affect safety~\cite{cp-nw-cui2022coopernaut,safety-nw-ernst2023application}). Optimizing application outcomes therefore requires system--AI co-design: reasoning jointly about models, placement, and networks, rather than any single component in isolation.

Existing simulation platforms address only fragments of this challenge. Pylot provides modular autonomy pipelines but assumes single-vehicle execution without cooperation. OpenCDA~\cite{xu2021opencda}, OpenCAMS~\cite{ahmad2025opencams}, and Plexe~\cite{segata2014plexe} incorporate aspects of V2X and network modeling, but do not capture the holistic interaction of autonomy stacks with heterogeneous computing infrastructure. As a result, critical questions remain unanswered: How does network jitter affect cooperative perception when the AV pipeline is split across the RSU and the ego vehicle? Does safety degradation occur when edge or cloud servers incorporated in the loop face workload spikes? How should perception tasks be redistributed between vehicles and infrastructure as network conditions fluctuate? No platform currently enables systematic exploration of these systems-autonomy interactions that determine real-world cooperative driving safety.

\begin{table*}[t!]\label{rel-work}
\centering
\renewcommand{\arraystretch}{1.3}
\setlength{\tabcolsep}{6pt}
\begin{tabular}{l
                >{\centering\arraybackslash}m{1.2cm}
                >{\centering\arraybackslash}m{2.0cm}
                >{\centering\arraybackslash}m{2.0cm}
                >{\centering\arraybackslash}m{2.0cm}
                >{\centering\arraybackslash}m{2.3cm}
                >{\centering\arraybackslash}m{1.5cm}}
\textbf{Platform} &
\textbf{V2X} &
\makecell{\textbf{Platform}\\\textbf{Heterogeneity}} &
\makecell{\textbf{Network}\\\textbf{Emulation}} &
\textbf{Orchestration} &
\makecell{\textbf{System-Level}\\\textbf{Metrics}} &
\textbf{Modularity} \\
\hline
Plexe~\cite{segata2014plexe}          & \fullcircle & \emptycircle & \halfcircle & \emptycircle & \emptycircle & \halfcircle \\
% \hline
Eclipse MOSAIC~\cite{schrab2022modeling}  & \fullcircle & \emptycircle & \fullcircle & \halfcircle  & \halfcircle  & \fullcircle \\
% \hline
V2XVerse~\cite{liu2025towards}        & \fullcircle & \emptycircle & \halfcircle & \emptycircle & \emptycircle & \fullcircle \\
% \hline
OpenCAMS~\cite{ahmad2025opencams}        & \fullcircle & \emptycircle & \fullcircle & \halfcircle  & \emptycircle & \fullcircle \\
% \hline
OpenCDA~\cite{xu2021opencda}         & \fullcircle & \emptycircle & \halfcircle & \halfcircle  & \emptycircle & \fullcircle \\
% \hline
VaN3Twin~\cite{pegurri2025van3twin}        & \fullcircle & \emptycircle & \fullcircle & \emptycircle & \emptycircle & \halfcircle \\
% \hline
\textbf{CADET [ours]}          & \redcircle  & \redcircle   & \redcircle  & \redcircle   & \redcircle   & \redcircle \\

\end{tabular}
\label{rel-work}
\caption{Comparison of capabilities across V2X simulation platforms. CADET uniquely integrates platform heterogeneity, orchestration, system-level metrics, and modularity, distinguishing it from prior simulators.}
\vspace{-1.2em}
\label{rel-work}
\end{table*}

This paper introduces CADET (Cooperative Autonomy through Distributed Experimentation Toolkit), a modular platform for reproducible evaluation of cooperative autonomy under realistic distributed systems constraints. Unlike monolithic AV stacks that execute fixed pipelines on single platforms, CADET disaggregates perception, planning, and control into composable modules deployable across vehicles, RSUs, edge nodes, and cloud servers. This architecture enables systematic exploration of distributed execution strategies and their impact on cooperative driving safety in V2V, V2I, and broader V2X scenarios.

CADET provides four core capabilities. First, a modular AV pipeline architecture supports flexible deployment of state-of-the-art models (YOLO~\cite{yolo11_ultralytics}, MPC~\cite{mpc}) and cooperation policies. Second, NetWaggle, a network emulation layer built on Mininet~\cite{fontes2015mininet}, injects realistic network dynamics modeling 4G/5G/WiFi variability and replayable traces. Third, distributed deployment abstractions enable seamless execution across heterogeneous resources, from resource-constrained vehicles to GPU clusters. Fourth, CADET incorporates an orchestration layer that, through a single configuration file, transparently manages model and policy selection, deployment, and network configuration. It further integrates multi-level instrumentation and records model-level (precision, recall, inference latency), system-level (queuing delay, bandwidth utilization, energy consumption), and task-level (time-to-collision, braking smoothness) metrics. Using such metrics, CADET makes it possible to trace how specific model, deployment, and network decisions translate into downstream application outcomes. Importantly, these components can be used individually or jointly, allowing not only the AV community, but also systems and ML researchers to evaluate their models in realistic cooperative settings.

In summary, this work makes the following contributions:
\begin{itemize}
\item We present CADET, the first modular platform that disaggregates the AV stack and supports distributed cooperative autonomy experiments with hardware-in-the-loop validation.
\item We develop NetWaggle, a reusable emulation layer that captures realistic V2X delay distributions and device heterogeneity.
\item We demonstrate CADET’s ability to compare cooperative autonomy policies, showing that V2V intent outperforms cloud-hosted perception under adverse network tails, and that RSU-assisted perception improves safety until compute saturation.
\item We release CADET as open-source software with documentation, benchmarks, and reference implementations to accelerate research at the intersection of distributed systems and cooperative autonomy.
\end{itemize}

\section{Related Work}

\subsection{Cooperative Autonomy}
Cooperative autonomy has emerged as a systems-level paradigm in which vehicles and infrastructure share state, intent, and sensor data to overcome the inherent limits of single-agent autonomy. Isolated AVs remain constrained by occlusion, restricted sensor range, and reaction delays~\cite{milanes2014modeling}, whereas vehicle-to-vehicle (V2V), vehicle-to-infrastructure (V2I), and broader vehicle-to-everything (V2X) cooperation extend perception and coordination. V2V enables cooperative adaptive cruise control (CACC) and platooning, yielding 10--30\% fuel savings~\cite{vanarem2006impact, milanes2014modeling}. Studies show that connected intersection managers can achieve 2--3$\times$ increase in intersection capacity and 90\% reduction in  delays~\cite{varaiya2016platoons}. 

However, these advantages hinge critically on the behavior of the underlying network and computational substrate. Empirical evaluations consistently show that latency, asynchrony, and packet loss degrade cooperative perception accuracy, with a sharp drop beyond 100 ms~\cite{nguyen2024v2xcp, elbatt2006cooperative}. Algorithms such as V2X-ViT try to address such challenges, yet temporal mismatches from communication delays continue to compromise perception outcomes~\cite{xu2022v2xvit}. Heterogeneous sensing and communication stacks exacerbate this, requiring bandwidth-aware scheduling and selective transmission to sustain performance under load~\cite{sharma2023impact, sharma2025towards, sharma2025cloud}. 

% Moreover, reliance on cloud infrastructure introduces variability from multi-tenancy and hidden prioritization policies that create non-deterministic behavior ~\cite{zhang2023edge}, challenging reproducibility and robustness.

\begin{figure*}[t!]
  \centering
  \begin{subfigure}{0.5\linewidth}
    \centering
    \includegraphics[scale=0.21]{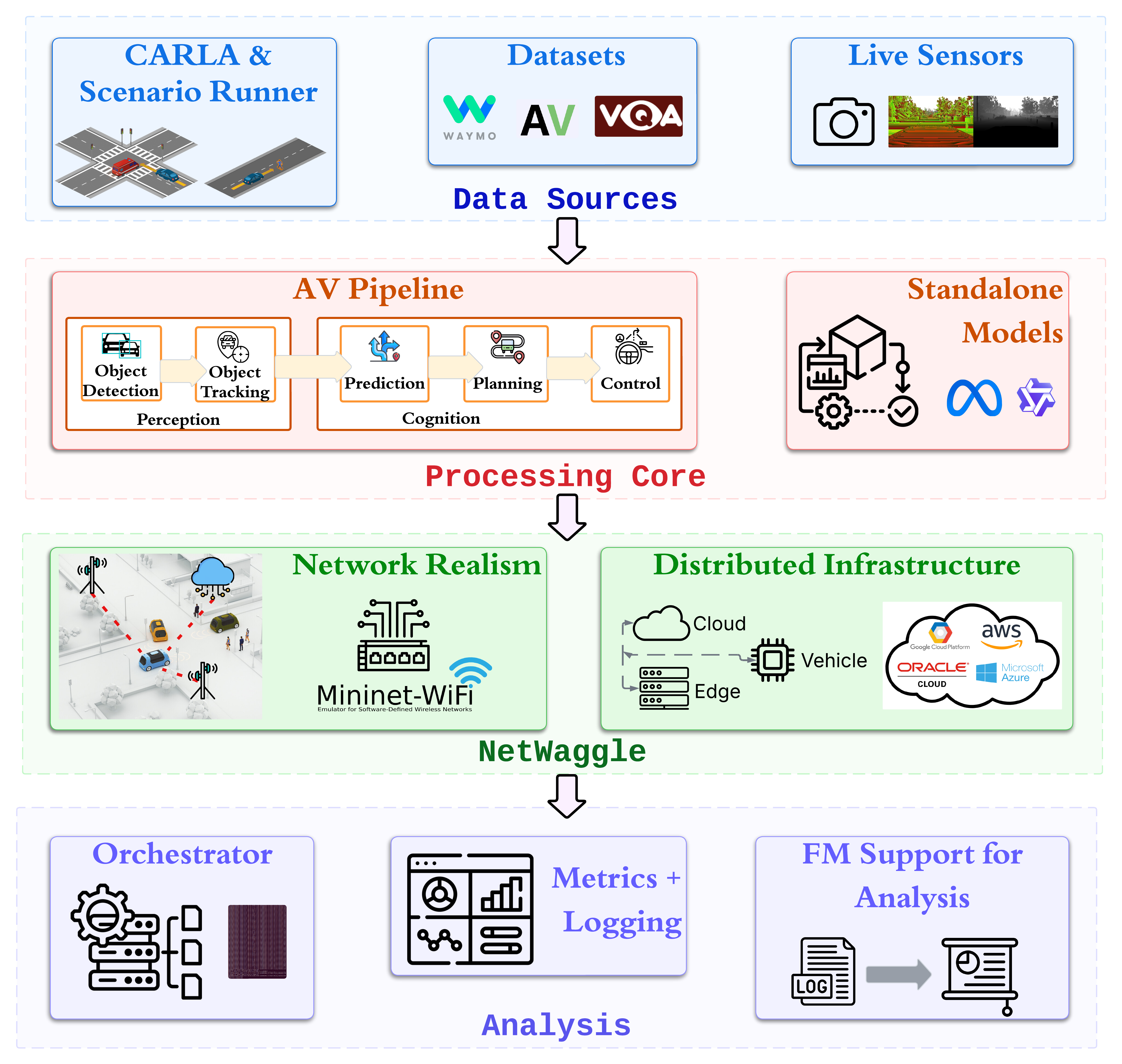}
    % \caption{CADET implements a modular four-layer architecture allowing components to be evaluated in isolation or integrated for end-to-end cooperative autonomy experiments.}
    \label{fig:architecture}
  \end{subfigure}\hfill
  \begin{subfigure}{0.45\linewidth}
    \centering
    \includegraphics[scale=0.28]{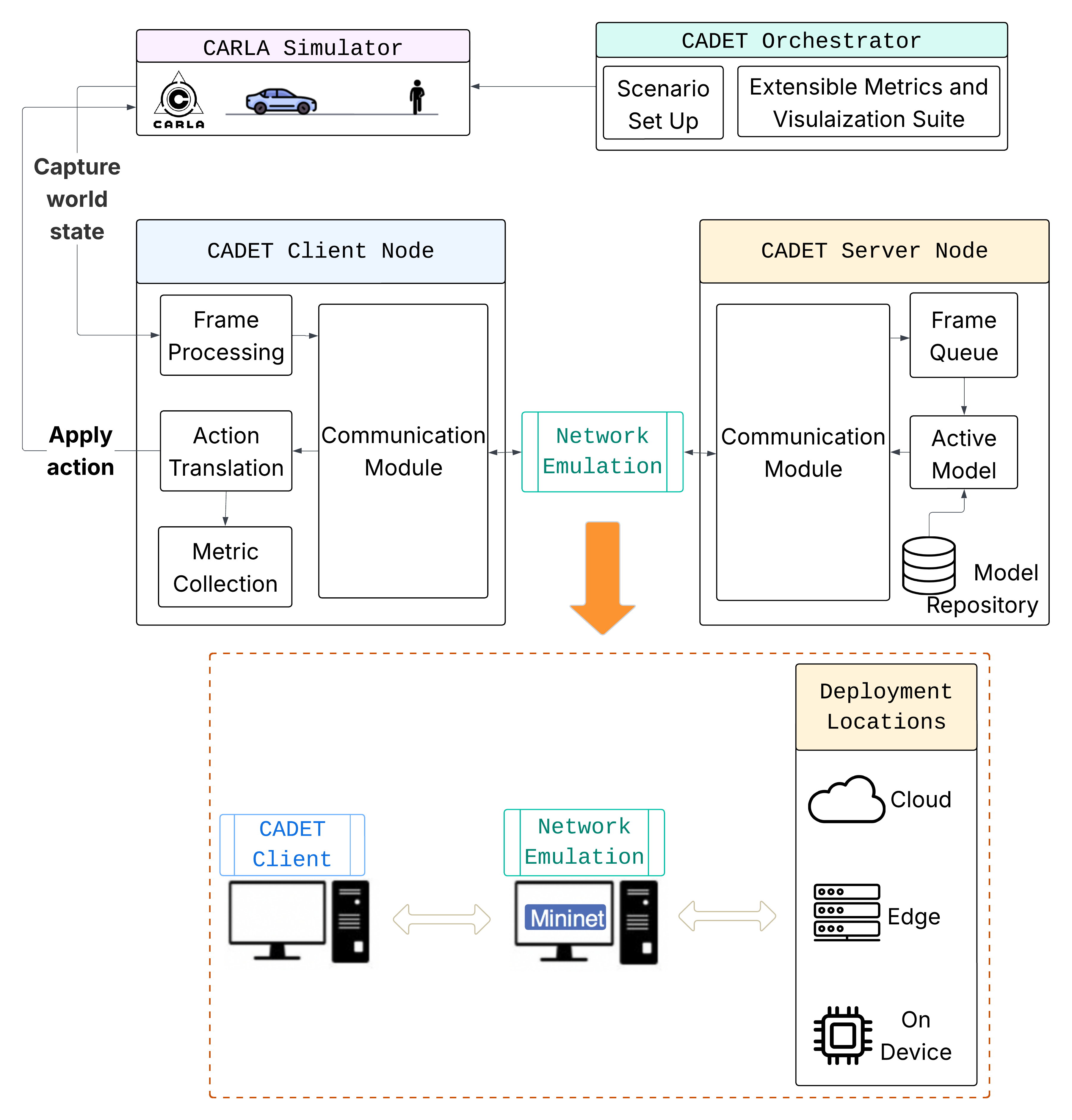}
    % \caption{NetWaggle enables distributed inference under device heterogeneity and realistic network dynamics. It adopts a client–server design and mimics real-world network profiles.}
    \label{fig:client-server}
  \end{subfigure}
  \vspace{-1em}
  \caption{\textbf{(left)} (a) CADET implements a modular four-layer architecture allowing components to be evaluated in isolation or integrated for end-to-end cooperative autonomy experiments. \textbf{(right)} (b) NetWaggle enables distributed inference under device heterogeneity and realistic network dynamics. It adopts a client–server design and mimics real-world network profiles.}
  \label{fig:two-figs}
  \vspace{-1em}
\end{figure*}

\subsection{V2X Simulation Platforms}

A diverse ecosystem of V2X simulation platforms has emerged to enable cooperative autonomy. Network-centric simulators such as Veins~\cite{sommer2019veins} pioneered SUMO-OMNeT++ coupling, while Plexe~\cite{segata2014plexe} added realistic vehicle dynamics for platooning and VaN3Twin~\cite{pegurri2025van3twin} integrated ray tracing to achieve 70\% accuracy improvements. Multi-domain platforms such as Eclipse MOSAIC~\cite{schrab2022modeling} enable runtime exchange across traffic, communication, and application domains, while OpenCAMS~\cite{ahmad2025opencams} synchronizes SUMO-CARLA-OMNeT++. Cooperative Driving Automation systems including V2XVerse~\cite{liu2025towards} demonstrate 62\% driving improvements through collaborative perception, and OpenCDA~\cite{xu2021opencda} provides Python pipelines for rapid prototyping. These platforms excel within their domains, achieving high-fidelity network modeling and realistic traffic dynamics.

However, existing simulators converge on a critical limitation: they overlook cooperative autonomy as a distributed systems problem that requires compute platform heterogeneity modeling and system-level orchestration. They fail to capture how AV pipelines are partitioned across heterogeneous compute tiers (e.g., perception on edge GPUs, planning on vehicle), each with distinct performance profiles. Additionally, real deployments face scenarios where multiple vehicles compete for limited cloud GPU cycles and processing rates drop from 30 FPS on datacenter hardware to 5 FPS on embedded accelerators. CADET addresses this gap (shown in Table~\ref{rel-work}) by providing multi-fidelity simulation and instrumentation that explicitly models computational resources across devices, hardware-in-the-loop testing for real deployment validation, and orchestration-aware metrics that capture end-to-end performance across distributed pipelines. 

\section{Design Goals} \label{des-goal}

The central goal of CADET is to enable systematic evaluation of cooperative autonomy algorithms under realistic V2X communication and computational constraints. To achieve this objective, CADET must fulfill four key requirements: 

\textbf{Modularity.} Cooperative autonomy research spans a wide spectrum of pipelines and scenarios, and CADET must support modularity at both the pipeline and component level. Researchers should be able to execute complete AV pipelines for end-to-end evaluation or isolate individual models (such as perception) for benchmarking in controlled settings. This flexibility allows components to be interchanged or assessed independently without re-implementation of the entire stack.

\textbf{Reproducibility.} Experiments involving distributed computation and network dynamics require careful orchestration to ensure deterministic outcomes. CADET must ensure that identical network, platform, and scenario configurations yield consistent results across repeated runs, enabling validation and comparison across users.

\textbf{Heterogeneity.} Cooperative autonomy operates across asymmetric sensing and computational resources, from onboard processors to edge/cloud servers. CADET should capture the performance profiles of this continuum while also integrating diverse data sources into a unified workflow.

\textbf{Multi-level Analysis.} Evaluating cooperative autonomy requires visibility across multiple layers of the system. CADET must provide synchronized instrumentation spanning model-level metrics, system-level behavior, and application-level outcomes. This fine-grained view enables researchers to attribute the observed performance to specific algorithmic, communication, or resource-level factors.

\section{CADET Architecture}\label{arch}
CADET implements a four-layer architecture that systematically addresses the design goals (Sec.~\ref{des-goal}) while maintaining the flexibility required for diverse cooperative autonomy research scenarios. The stratified design, shown in Fig.~\ref{fig:two-figs}(a) establishes clear abstraction boundaries that facilitate both isolated component testing and integrated system evaluation.

\subsection{Data Source Layer}
The \texttt{Data Source Layer} provides scenario generation and sensing inputs integration via three interfaces: CARLA simulator, datasets, and real-world sensor inputs.  

At the core of this layer is \texttt{CARLA}~\cite{dosovitskiy2017carla}, an open-source simulator widely adopted in the AV community. CARLA couples a physics-based vehicle dynamics engine with sensor-accurate rendering, enabling controlled evaluation of perception, planning, and control algorithms in urban environments. Augmented with the \texttt{Scenario Runner} library, it supports configurable V2X scenarios such as cooperative perception and platooning maneuvers. Researchers can instantiate road networks, spawn vehicles or pedestrians with scripted behaviors, and configure environmental conditions (e.g., rain, night driving, occlusion). Integration with Scenario Runner ensures modularity: new actors, pre-defined or cooperative behaviors can be introduced without altering the core simulation. This framework also supports failure-mode injection, such as dynamic changes in actors' behaviors or RSU disconnections, providing realistic stressors for cooperative autonomy scenarios.  

% For instance, researchers can simulate a rainy-night pedestrian crossing at an urban intersection while configuring different cooperative perception modules across distributed vehicles.

% In addition to live simulation, CADET’s \texttt{Dataset Mode} allows experiments to run on pre-recorded sensor inputs and benchmarking datasets. This enables deterministic replay, which is essential for controlled comparison of cooperative perception and fusion strategies under varying network conditions. Dataset-based evaluation maintains compatibility with downstream components, including network emulation and the metrics framework, ensuring reproducibility even without real-time execution. While primarily designed for AV benchmarks such as Waymo Open Dataset~\cite{sun2020scalability-waymo}, CADET can also incorporate datasets beyond AV contexts (e.g., COCO~\cite{lin2014coco} for object detection).

CADET's \texttt{Dataset Mode} enables deterministic replay on pre-recorded inputs (e.g., Waymo~\cite{sun2020scalability-waymo}, COCO~\cite{lin2014coco}), maintaining full compatibility with network emulation and metrics collection for reproducible evaluation without real-time simulation. Finally, the data source layer can incorporate \texttt{Live Sensor} inputs, enabling direct evaluation with physical hardware (e.g., RGB camera, LiDAR) or hybrid setups that combine real and simulated streams. 

\subsection{Processing Core Layer}

The \texttt{Processing Core Layer} embodies the modularity design goal through its configurable AV pipeline architecture and standalone model interfaces.  

The \texttt{AV Pipeline} is decomposed into two primary stages: Perception and Cognition, each comprising multiple sub-stages that can be independently configured, swapped, and deployed across the computational continuum. For instance, object detection may be executed on a cloud server with high compute capacity, while trajectory planning runs on an RSU at the edge.  

The Perception stage integrates object detection, tracking, and localization across local and V2V-shared observations, fusing asynchronous detections from multiple agents. The Cognition stage uses this enhanced awareness for prediction, planning, and control, enabling more accurate trajectory anticipation and safer motion plans.

Beyond the full AV pipeline, CADET provides an interface for evaluating \texttt{Standalone Models} in isolation. Users can benchmark perception, prediction, or emerging foundation models~\cite{bai2023qwen, touvron2023llama} without running the entire AV pipeline, enabling fine-grained analysis of how network-induced delays or resource constraints affect specific algorithms.

\begin{figure}[t]
    \setlength{\abovecaptionskip}{0.2cm}
    \centerline{\includegraphics[scale=0.3]{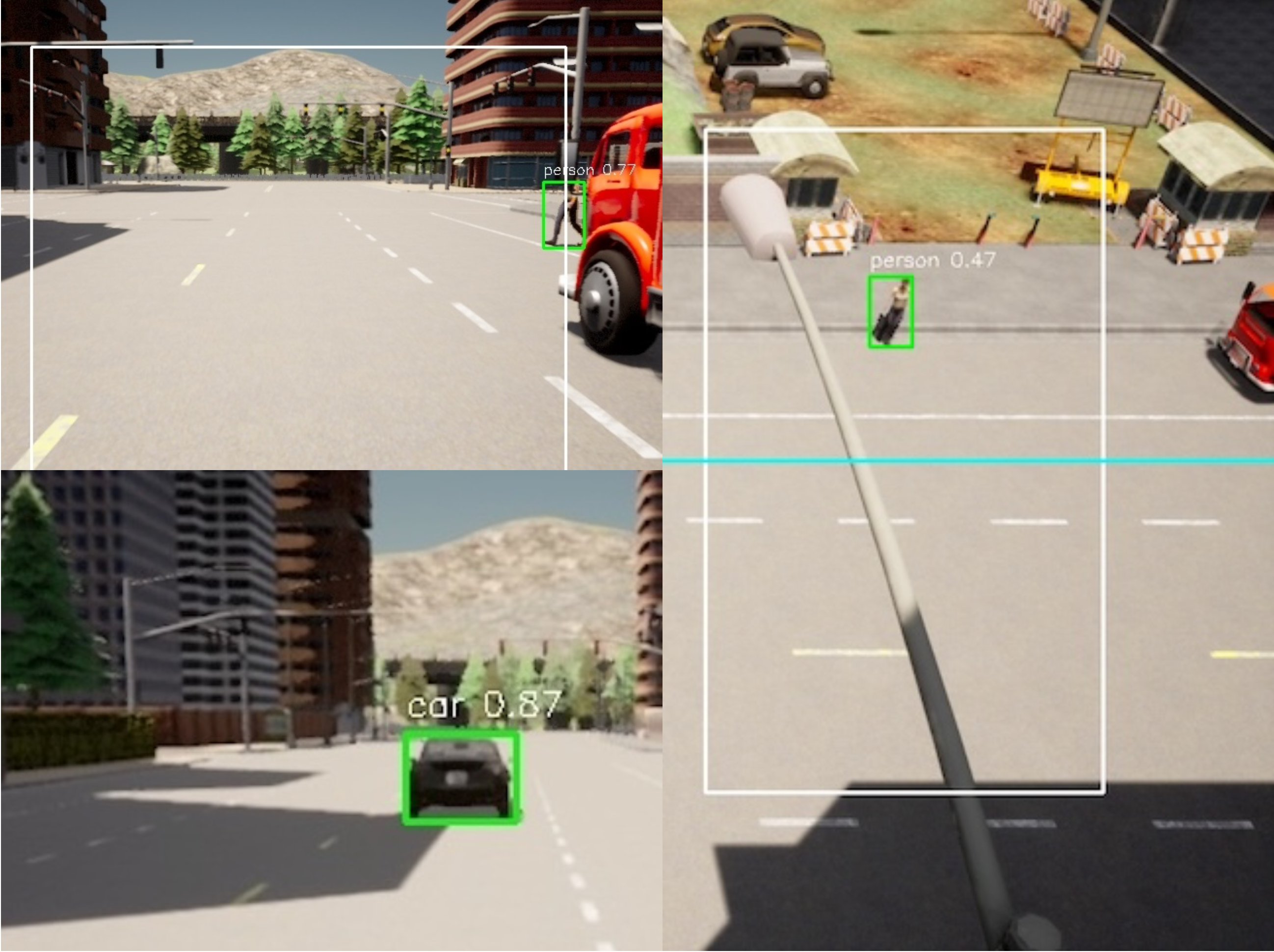}}
    \centering
    \caption{Example CADET simulation captures: (a) \textbf{(bottom-left)} V2V scenario showing a platooning leader–follower interaction; (b) \textbf{(top-left)} ego vehicle camera view. \textbf{(right)} infrastructure camera view illustrating V2I perception.}
    \label{fig:sim}
    \vspace{-1.5em}
\end{figure}

\subsection{NetWaggle}

Distributed cooperative autonomy requires components of the AV pipeline to execute across heterogeneous devices (onboard processors, roadside units (RSUs), edge servers, and cloud platforms) connected through real networks. While such deployments are essential for realism, they are notoriously difficult to evaluate reproducibly. Actual cloud and network environments introduce variability from multi-tenancy, time-of-day fluctuations, congestion, and hidden prioritization policies~\cite{sharma2025cloud}. These factors make it impossible to obtain deterministic outcomes: the same cooperative braking scenario might succeed in one run and fail in another purely due to transient network or resource conditions. To fulfill CADET's design goals of reproducibility and heterogeneity, we introduce \texttt{NetWaggle}, a network emulation layer that combines device heterogeneity with network realism in a controlled and repeatable setting.  

\textbf{Device heterogeneity.} NetWaggle supports deployment across a continuum of computational platforms, from embedded vehicle-like devices to GPU-equipped RSUs and cloud servers. To enable this flexibility, CADET adopts a client--server architecture: the client hosts the simulation environment and manages vehicle actuation, while servers execute perception, prediction, or planning modules. This separation, shown in Fig~\ref{fig:two-figs}(b), allows the AV pipeline to run entirely on the client, fully offloaded to the server, or distributed across multiple nodes. The architecture naturally extends to multi-vehicle scenarios, where each vehicle operates as a client while cooperating via shared infrastructure or peer-to-peer channels. To better reflect real deployments, CADET departs from CARLA's synchronous execution model and supports an asynchronous mode in which the simulation progresses even while awaiting delayed commands, mimicking the timing behavior of real-world V2X systems. Furthermore, CADET experiments can incorporate real devices, with modules deployed directly to local GPUs, Jetson boards, or cloud VMs, enabling hardware-in-the-loop testing alongside simulation. Data exchange between nodes is implemented over WebSockets~\cite{fette2011websocket}, supporting both frame-by-frame and streaming modes to match the requirements of different cooperative autonomy pipelines.  

\textbf{Network realism.} Alongside heterogeneous devices, cooperative autonomy depends critically on the properties of the connecting networks. Real-world V2X links are affected by network dynamics which influences safety outcomes~\cite{safety-nw-ernst2023application}. Using uncontrolled networks directly in experiments undermines reproducibility. NetWaggle addresses this challenge by interposing Mininet-based emulation between distributed components. Researchers can inject latency distributions, jitter patterns, packet loss rates, and bandwidth caps that match measured traces from 4G-LTE, 5G-NR, DSRC, or congested Wi-Fi links~\cite{coll2022end}, or synthetically construct profiles to stress-test cooperative behaviors under adverse conditions. Mininet's low overhead (typically $<$10 ms) ensures that emulation does not distort experimental outcomes while preserving repeatability. NetWaggle can connect both local heterogeneous devices and remote servers, enabling experiments that are simultaneously realistic and reproducible.  

% By unifying device heterogeneity and network realism under a single abstraction, NetWaggle provides researchers with a flexible substrate to evaluate cooperative autonomy. It enables systematic studies of how communication and computation trade-offs shape outcomes such as platoon stability~\cite{platooning_safety}, intersection coordination~\cite{intersection_coordination}, or RSU-assisted perception under degraded connectivity~\cite{cooperative_perception}. 

\subsection{Analysis Layer}

The \texttt{Analysis Layer} operationalizes evaluation by coordinating experiments, collecting metrics, and enabling reproducible analysis across distributed cooperative autonomy scenarios. It comprises three components:   

\texttt{Orchestrator.} The Orchestrator manages the full lifecycle of an experiment, unifying configuration and execution. A single configuration file specifies scenario parameters, model variants, deployment mappings (cloud, onboard), and network traces. The orchestrator parses this configuration and launches all components in sequence: CARLA with Scenario Runner, client nodes (ego and cooperative vehicles), and distributed inference servers. Synchronization ensures that V2X communication channels are active before scenario execution begins. Runtime orchestration supports remote deployment through SSH-based execution, allowing servers to run on heterogeneous devices or cloud resources while remaining under centralized control. Batch execution mode enables systematic sweeps over parameters such as vehicle density, network delay profiles, or concurrent workloads.  

\texttt{Metrics and Logging.} CADET provides synchronized metrics across three dimensions: (i) model-level metrics capturing algorithmic accuracy (e.g., cooperative object detection mAP, tracking consistency) and inference latency; (ii) performance outcomes measuring task-level behavior relevant to safety, including time-to-collision and braking efficiency; and (iii) system-level metrics recording CPU/GPU utilization, memory footprint, and energy consumption, characterizing the feasibility of deploying cooperative autonomy policies on diverse platforms. Logs are timestamped at critical points in the pipeline (frame capture, transmission, inference start/end, actuation) to attribute latency between computation, communication, and coordination overheads. Time synchronization libraries (e.g., NTP~\cite{mills2006computer}) align distributed logs across clients, RSUs, and servers, supporting consistent analysis of end-to-end behavior. The logger outputs structured records (CSV/JSON) that can feed both real-time dashboards and post-hoc statistical evaluation, facilitating reproducible benchmarking across experiments.  

\texttt{Foundation model interface.} As an extension, CADET can route logs through vision-language or language models~\cite{touvron2023llama} to provide semantic summarization and exploratory analysis. While not central to CADET's architecture, this demonstrates extensibility toward foundation-model-based cooperative autonomy.

\section{Evaluation}
%   % -------- Row 2 --------
%   \begin{subfigure}[t]{0.33\textwidth}
%     \centering
%     \includegraphics[width=\linewidth]{ieeeconf/figures/accuracy_latency_scatter_multi.png}
%     \caption{V2I: Throughput vs. Batch}
%     \label{fig:v2i-plot2}
%   \end{subfigure}\hfill
%   \begin{subfigure}[t]{0.33\textwidth}
%     \centering
%     \includegraphics[width=\linewidth]{ieeeconf/figures/latency_violin_plot.png}
%     \caption{Placeholder caption}
%     \label{fig:plot5}
%   \end{subfigure}\hfill
%   \begin{subfigure}[t]{0.25\textwidth}
%     \centering
%     \includegraphics[width=\linewidth]{ieeeconf/figures/sim_runs-2.png}
%     \caption{Placeholder caption}
%     \label{fig:plot6}
%   \end{subfigure}

%   \vspace{0.8em}

%   \caption{Composite 3×3 grid of plots across V2V, V2I, and DatasetMod experiments.}
%   \label{fig:plots-3x3}
% \end{figure*}

% \include{ieeeconf/figures/collision-heatmap}

% \include{ieeeconf/figures/dataset-table}

% \include{ieeeconf/figures/simulation_frames}

\begin{figure}[t]\label{fig:v2v-plot}
    \setlength{\abovecaptionskip}{0.2cm}
    \centerline{\includegraphics[scale=0.4]{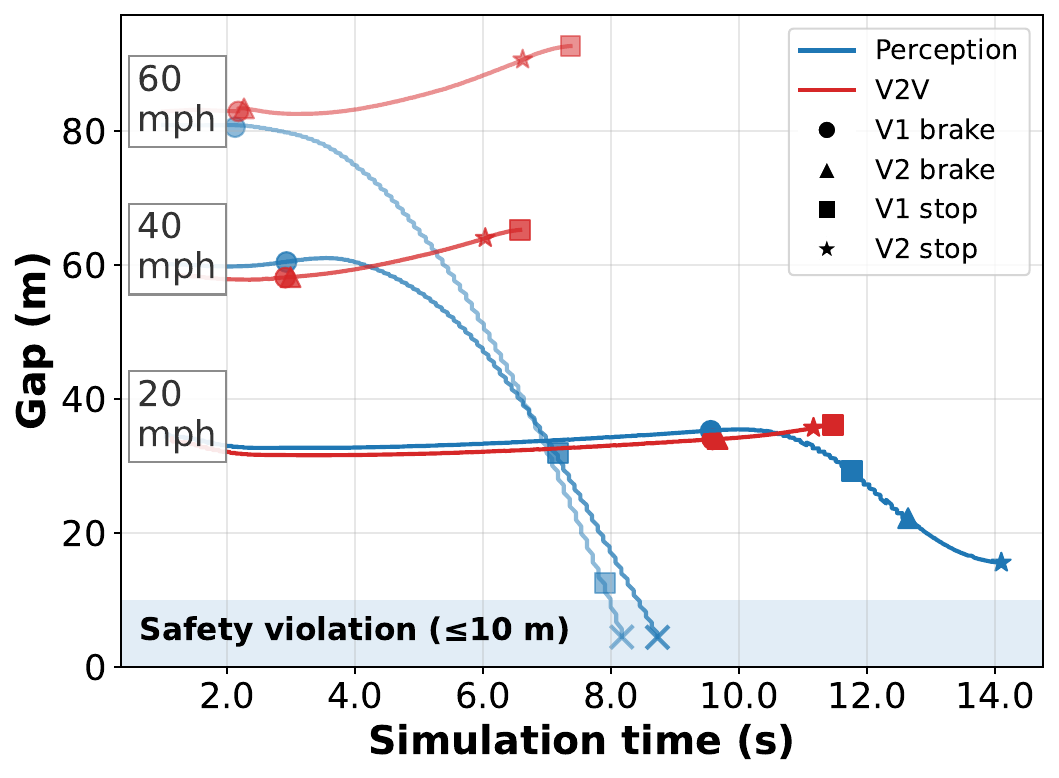}}
    \centering
    \caption{(V2V) Leader–follower gap over time at different speeds  under perception-only and V2V-based braking.}
    \label{fig:v2v-plot}
    \vspace{-1.5em}
\end{figure}

We evaluate CADET through experiments that validate its design goals and demonstrate its unique capability to reveal how systems-level decisions fundamentally alter cooperative autonomy outcomes. Our evaluation establishes that CADET transforms complex distributed experimentation, that traditionally require thousands of lines of specialized code, into concise, reproducible studies accessible through simple configuration files. Each experiment isolates specific dimensions of the cooperative autonomy design space (e.g., model architecture, deployment topology, concurrent load) to attribute safety outcomes to individual factors rather than aggregate system behavior.

\subsection{Implementation and Platform Validation}

CADET comprises $ \sim$30k lines of Python code implementing the architecture described in Sec.~\ref{arch}, with NetWaggle enabling distributed coordination across the vehicle-cloud continuum. Despite this implementation complexity, researchers interact with CADET through declarative YAML configuration files that specify scenario parameters, model deployments, and network profiles. For instance, a complete V2I perception-sharing experiment requires only 8-10 lines of code for scenario configuration.\footnote{Further implementation, configuration, and evaluation details can be found at https://nesl.github.io/cadet-web}

To validate CADET’s modularity and reproducibility, we repeated identical cooperative autonomy scenarios 20 times under a fixed deployment configuration and network profile. Across repetitions, we observed $<$$5$\% variance in safety metrics (e.g., time-to-collision and collision outcomes) and $<$$3$~ms standard deviation in end-to-end latency, confirming CADET’s ability to deliver reproducible results despite the inherent nuances of distributed execution.

We next demonstrate that CADET orchestrates cooperative autonomy scenarios, including V2V, V2I, and dataset-driven experiments, each designed to stress-test different dimensions of distributed autonomy.

\begin{figure*}[t]
  \centering

  % -------- Row 1 --------
  \begin{subfigure}[t]{0.35\textwidth}
    \centering
    \includegraphics[width=\linewidth]{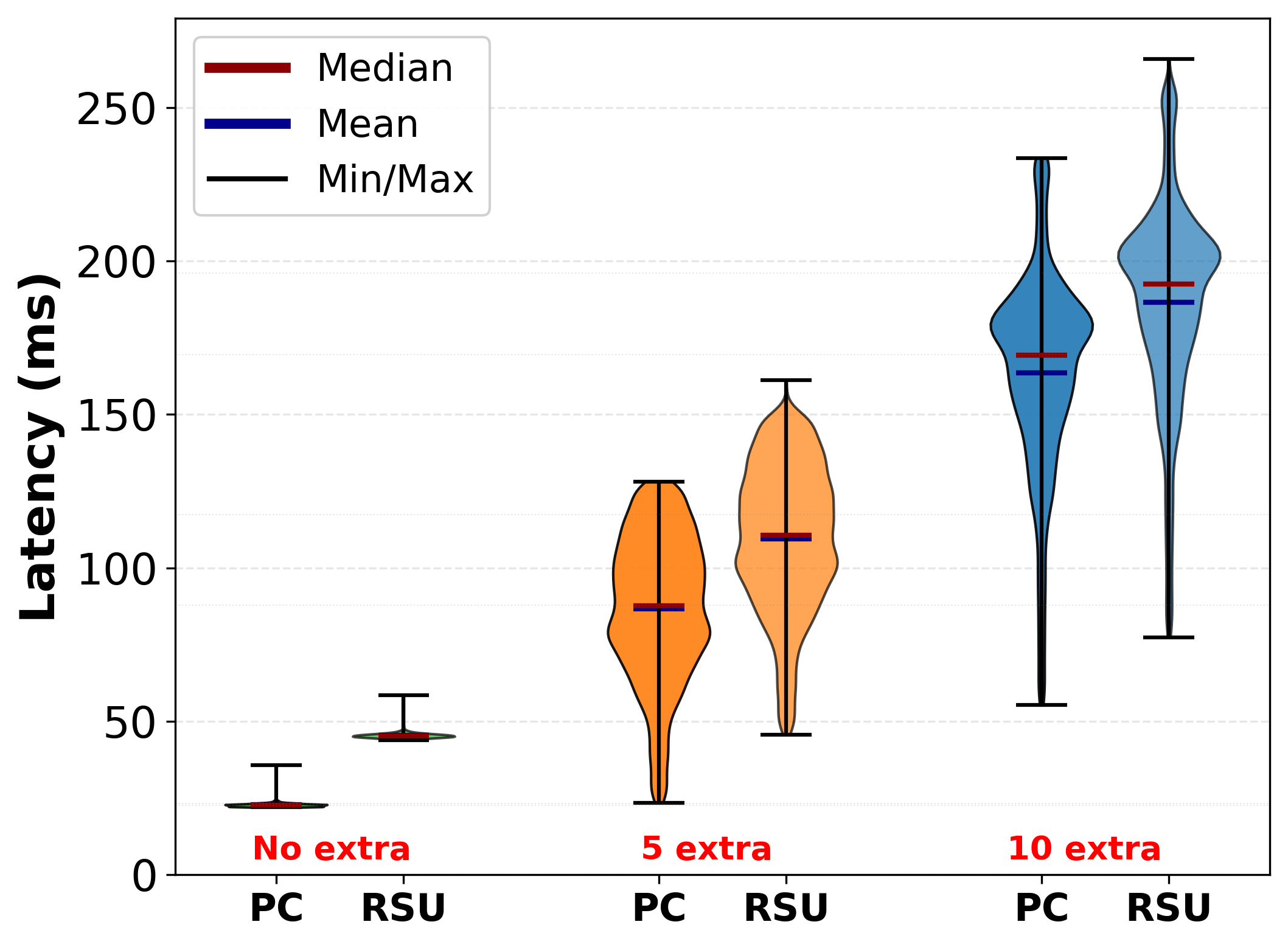}
    % \caption{V2V: Gap between leader (V1) and follower (V2)}
    % \label{fig:v2i}
  \end{subfigure}\hfill
  \begin{subfigure}[t]{0.6\textwidth}
    \centering
    \includegraphics[width=\linewidth]{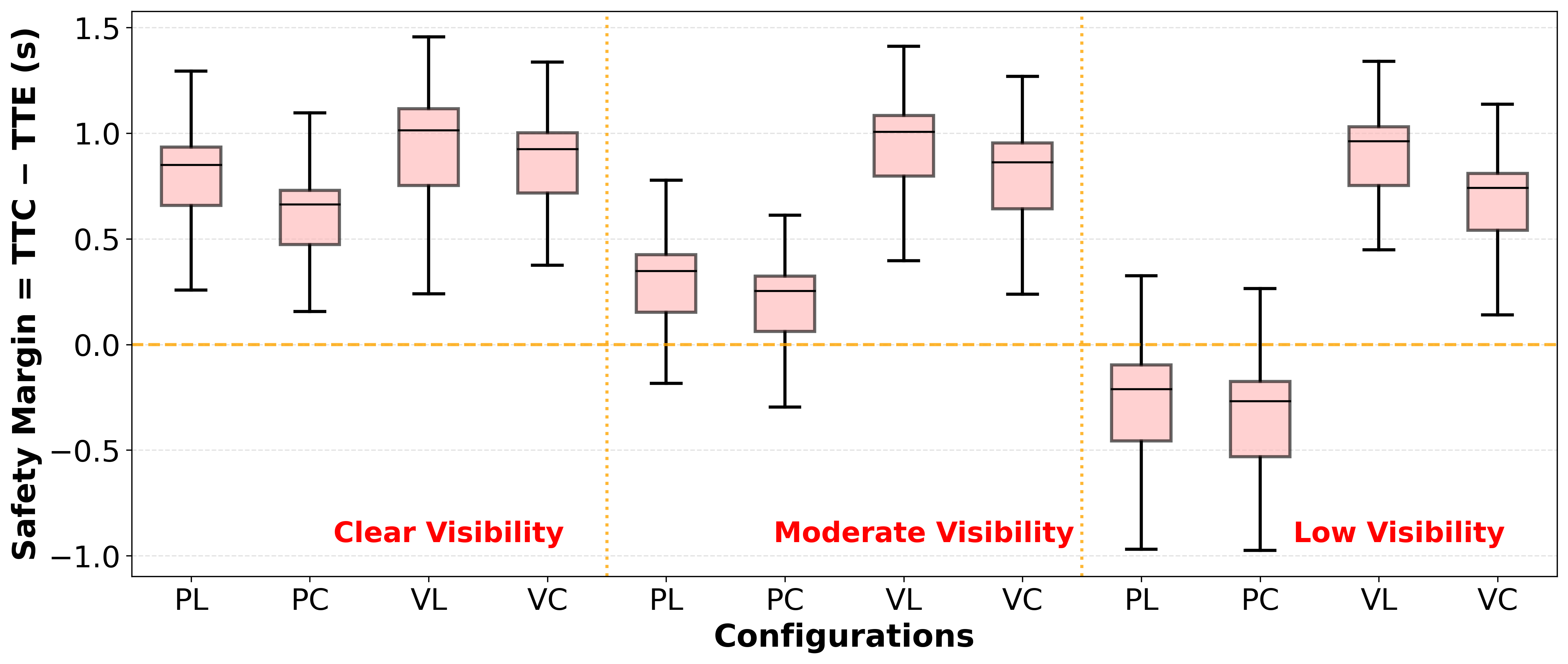}
    % \caption{V2I: Safety Margin at 20mph}
    % \label{fig:v2i-plot1}
  \end{subfigure}
\vspace{-0.2em}
  \caption{V2I evaluation results. \textbf{(left)} (a) End-to-end latency distribution under increasing concurrent load (no extra, +5, +10 clients) for perception on cloud (PC) and RSU. \textbf{(right)} (b) Safety margin across visibility conditions for different pipelines: PL (perception-only local), PC (perception-only cloud), VL (V2I local), and VC (V2I cloud). Results show stable positive margins in clear visibility, with margins shrinking and turning negative under occlusion.}
  \label{fig:v2i}
  \vspace{-1.3em}
\end{figure*}

\subsection{V2V Cooperative Braking}
We begin with a two-vehicle platoon scenario as a representative case of V2V safety-critical coordination (Fig.~\ref{fig:sim}(a)). The leader executes a single emergency braking event, and the follower must respond quickly enough to maintain a safe inter-vehicle distance. This scenario captures a fundamental trade-off in cooperative autonomy: whether safety is better served by explicit V2V intent packets or by high-capacity perception models operating under networked inference.  

\textbf{Experimental Setup.} The safety threshold for this experiment is set to $10\,\text{m}$, approximately two car lengths. The follower operates under two modes: (i) perception-only, using cloud-hosted YOLO11x inference deployed on an datacenter-class GPU (NVIDIA A5000), and (ii) V2V intent communication, where the leader transmits a braking packet directly. Cloud-based perception is modeled with a $30\,\text{ms}$ baseline ($p50$) communication latency consistent with 5G connectivity, with delay distributions extending to $100\,\text{ms}$ at the tail ($p99$). V2V packets are modeled after DSRC links with a baseline of $20\,\text{ms}$ and a maximum tail of $50\,\text{ms}$. While Fig.~\ref{fig:v2v-plot} highlights YOLO11x and V2V for clarity, the full evaluation was run across all YOLO variants from nano to extra-large, with consistent patterns observed.  

\textbf{Results.} Fig.~\ref{fig:v2v-plot} plots the inter-vehicle gap across speeds of $20$, $40$, and $60\,\text{mph}$. At $20\,\text{mph}$, perception-only control maintains a safe margin, but at $40\,\text{mph}$ and $60\,\text{mph}$, the follower approaches within $10\,\text{m}$ of the leader even before braking, violating the safety threshold. In contrast, V2V policy maintains a substantially larger gap, with margins exceeding $12\,\text{m}$ even at higher speeds. The timing of brake initiation (denoted by \scalebox{0.65}{\ding{108}} and \scalebox{0.65}{\ding{115}}) confirms that V2V followers react nearly instantaneously to leader intent, whereas perception-only followers accumulate additional delay from frame transmission and cloud inference.  

Table~\ref{tab:v2v} further quantifies collision outcomes under varying network delay percentiles. Perception-only control fails at $p90$ and $p99$, with both larger models resulting in collisions despite their higher perception accuracy. The root cause is compounding delay: $30$--$50\,\text{ms}$ inference latencies added to $100\,\text{ms}$ tail delays exceed the safe response window. Moreover, larger models impose higher energy costs (up to $1.66\,\text{J}$ per inference) without improving safety under adverse network conditions. In contrast, V2V communication remains collision-free across all delay profiles, as intent packets are small and efficient to transmit.

\textbf{CADET's Impact.} This experiment illustrates how CADET operationalizes reproducible evaluation of cooperative autonomy. With a single configuration file, CADET automatically records fine-grained timing traces, such as braking onset, stopping distance, and inter-vehicle gap, while simultaneously capturing system-level metrics including per-inference energy. 
% This integration enables systematic stress-testing across speeds, network delay profiles, and perception model scales without manual re-implementation.  

\begin{table}[t]\label{tab:v2v}
\centering
\label{tab:v2v}
\renewcommand{\arraystretch}{1.2}
\setlength{\tabcolsep}{3.5pt}
\arrayrulecolor{black}
\setlength{\arrayrulewidth}{0.4pt}
\label{tab:v2v}
% ----- Subtable A -----
\begin{subtable}{\linewidth}
\centering
\begin{tabular}{>{\centering\arraybackslash}p{1.2cm}|
                >{\centering\arraybackslash}p{1.2cm}|
                >{\centering\arraybackslash}p{1.2cm}|
                c|c|c|c}
\hline
\multirow{2}{*}{\textbf{Policy}}
& \multirow{2}{*}{\makecell{\textbf{Object}\\\textbf{Detector}}}
& \multirow{2}{*}{\makecell{\textbf{Energy}\\\textbf{(J)}}}
& \multicolumn{4}{c}{\textbf{Network Delays}} \\
\cline{4-7}
& & & \textbf{p50} & \textbf{p80} & \textbf{p90} & \textbf{p99} \\
\hline
V2V & -- & -- & \cellcolor{green!80} & \cellcolor{green!80} & \cellcolor{green!80} & \cellcolor{green!80} \\
\hline
\multirow{5}{*}{\begin{tabular}[c]{@{}c@{}}Perception\\(Cloud)\end{tabular}}
& YOLO11n & 0.73 & \cellcolor{green!80} & \cellcolor{green!80} & \cellcolor{green!80} & \cellcolor{green!80} \\
\cline{2-7}
& YOLO11s & 0.76 & \cellcolor{green!80} & \cellcolor{green!80} & \cellcolor{green!80} & \cellcolor{green!80} \\
\cline{2-7}
& YOLO11m & 0.92 & \cellcolor{green!80} & \cellcolor{green!80} & \cellcolor{green!80} & \cellcolor{green!80} \\
\cline{2-7}
& YOLO11l & 1.27 & \cellcolor{green!80} & \cellcolor{green!80} & \cellcolor{red!80} & \cellcolor{red!80} \\
\cline{2-7}
& YOLO11x & 1.66 & \cellcolor{green!80} & \cellcolor{green!80} & \cellcolor{red!80} & \cellcolor{red!80} \\
\hline
\end{tabular}
\centering
% \vspace{0.5em}
% \caption{Policies vs. network delays}
\end{subtable}
\caption{V2V Collision outcomes across various deployment configurations.  (\color{red!80}{Red} \color{black}: Collision, \color{green!80}{Green} \color{black}: No Collision)}
\vspace{-1.5em}
\label{tab:v2v}
\end{table}

\subsection{V2I-Assisted Braking}

The second demonstration evaluates infrastructure-assisted perception in a braking scenario where roadside units extend the sensing capabilities of the ego vehicle beyond its line of sight, as shown in Fig.~\ref{fig:sim}(b).

% This case captures the core promise of V2I cooperation: enabling timely reactions to occluded hazards that the vehicle cannot perceive directly.  

\textbf{Experimental Setup.} An ego vehicle travels at $20\,\text{mph}$ when a pedestrian emerges from behind an occluding truck and the ego must brake in time to avoid a collision. We define the safety margin as the difference between the ego vehicle’s time to reach the collision area (TTC) and the pedestrian’s time to enter it (TTE), quantifying the temporal buffer available at detection. We use it as a post-hoc measure of timing safety, indicating whether detection and braking occurred with sufficient buffer to avoid overlap. We evaluate four configurations: (i) onboard perception using YOLO11x, (ii) cloud-hosted perception with additional $30\,\text{ms}$ network delay, (iii) RSU-local V2I, where an infrastructure-mounted camera is co-located with the RSU and inference is performed locally (no additional communication latency), and (iv) RSU-cloud V2I, where the infrastructure camera is networked with the RSU, incurring an extra $30\,\text{ms}$ hop before inference. Visibility conditions vary from clear to moderate to low, simulating weather-related sensor degradation. We additionally examine scalability by adding concurrent clients (simulating additional load by 5-10 vehicles) generating requests to the RSU and Cloud server.  

\textbf{Results.} Fig.~\ref{fig:v2i}(b) shows that perception-only policies achieve sufficient safety margins under clear visibility but deteriorates rapidly as conditions worsen; under low visibility, the median safety margin drops below zero, indicating unsafe operation even at $20\,\text{mph}$. In contrast, both RSU-local and RSU-cloud V2I configurations maintain positive margins across all visibility settings, demonstrating the advantage of infrastructure-based sensing for occlusion handling. Additionally, Figure~\ref{fig:v2i}(a) shows RSUs exceed $100\,\text{ms}$ inference latency with 5-10 extra clients, and degrade gradually as load increases, becoming compute-bound beyond 10 vehicles. Cloud-based deployments, while elastically scalable, still exhibit long-tailed latency distributions: $p99$ latencies exceed $200\,\text{ms}$ with 10 clients due to queuing under concurrent load. These results reveal complementary failure modes: perception-only policies fail under degraded visibility, while V2I policies fail under heavy compute load.  

\textbf{CADET's Impact.} This experiment highlights CADET’s ability to couple safety evaluation with scalability analysis in a single framework. CADET orchestrated multi-node deployments spanning vehicle, RSU, and cloud, while injecting additional load profiles.

\subsection{Dataset Mode with Foundation Models}

Our final demonstration uses CADET’s dataset mode to explore the feasibility of integrating vision-language models (VLMs) into cooperative autonomy pipelines. While current cooperative autonomy relies on specialized DNNs for perception and control, the growing capability of VLMs suggests future interest in semantic V2X communication, where vehicles or infrastructure may exchange natural language descriptions of complex scenes.  

\begin{figure}[t]
    \setlength{\abovecaptionskip}{0cm}
    \centerline{\includegraphics[scale=0.4]{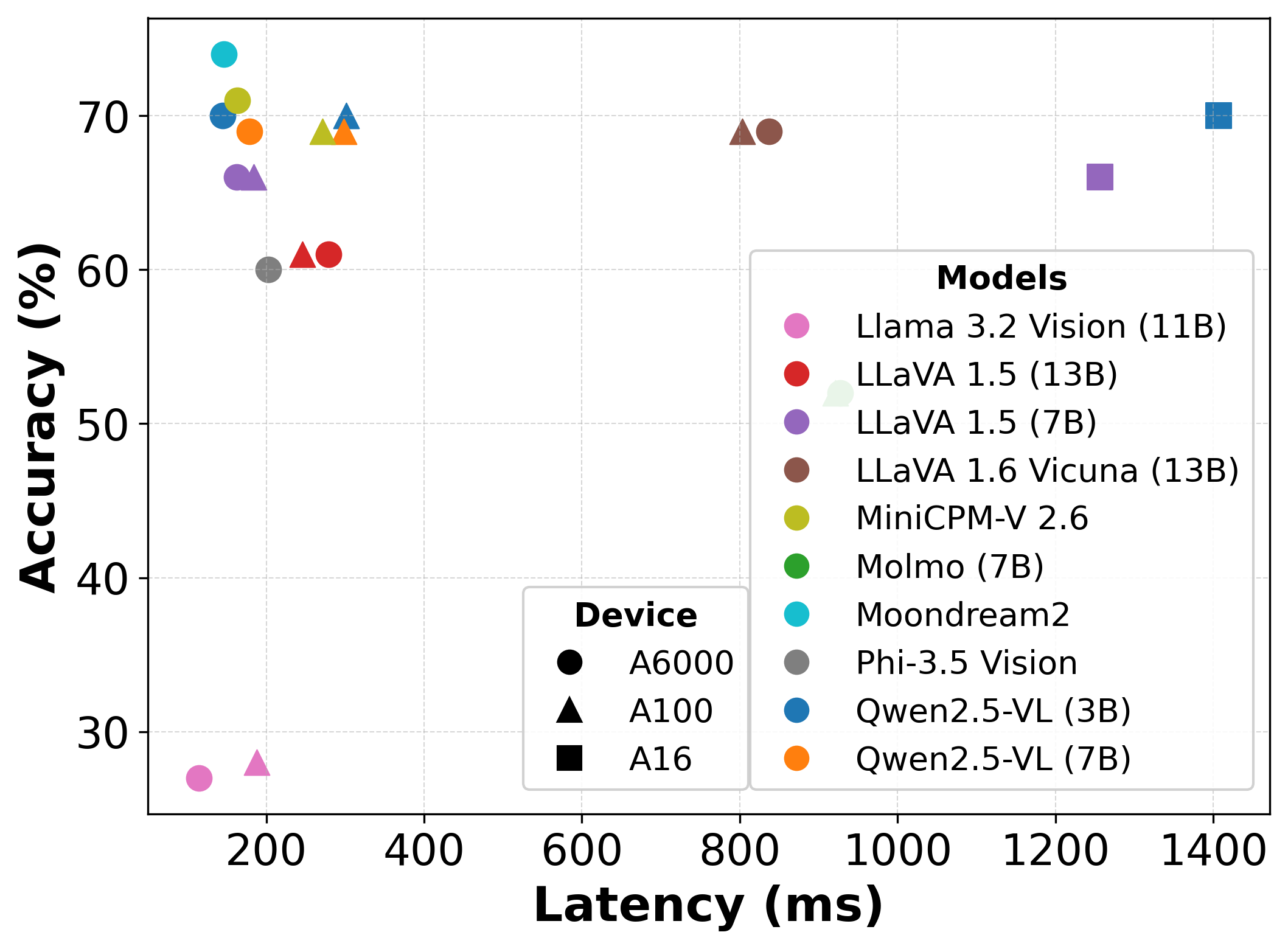}}
    \centering
    \caption{Accuracy–latency tradeoffs for different VLMs across heterogeneous deployment platforms.}
    \label{fig:dataset}
    \vspace{-1.5em}
\end{figure}

\textbf{Experimental Setup and Results.} We evaluate ten VLMs across three GPU servers (NVIDIA A6000, A100, and A16) using the VQA2 dataset. This evaluation serves as a proxy for cooperative autonomy scenarios where sensor or scene data is sent to a foundation model and queried in natural language, rather than being decomposed into narrowly scoped perception or planning tasks. Fig.~\ref{fig:dataset} shows non-intuitive accuracy-latency profile that do not seem to improve with increasing model size or number of parameters. Further, inference latencies exceeding $200$ms even for smaller models lie far beyond the deadlines of safety-critical coordination. We also measure the corresponding resource demands: 13B-scale models require more than $25\,\text{GB}$ of GPU memory and deliver throughput below $10$ tokens/s, while some 3B--7B models remain efficient, sustaining $>$$12$ tokens/s with manageable memory footprints. These findings indicate that, although semantic queries are promising for higher-level coordination or advisory tasks, current models are too slow and resource-intensive for low-level safety decisions.  

\textbf{CADET's Impact.} By enabling dataset-driven evaluation, CADET makes it possible to benchmark standalone models such as LLMs and VLMs alongside conventional DNNs using the same unified metrics and infrastructure. As these models continue to scale, CADET provides the infrastructure to systematically test when, where, and how such models could augment cooperative autonomy.

\section{Conclusion and Discussion}

 This work introduced CADET, a modular platform that disaggregates the AV stack into composable modules deployable across the vehicle–edge–cloud continuum. By coupling modular pipelines with NetWaggle’s reproducible network emulation and unified orchestration, CADET enables controlled evaluation of cooperative autonomy under heterogeneous deployment conditions. Our experiments show that safety outcomes hinge as much on \textit{where} and \textit{how} models are executed as on their algorithmic accuracy. V2V braking scenarios reveal that low-latency intent packets consistently outperform cloud-based perception despite the latter’s higher fidelity, while V2I experiments highlight that RSU-assisted perception preserves positive safety margins under occlusion yet collapses when overloaded by concurrent clients. Dataset-driven evaluations further demonstrate that large vision–language models, though attractive for semantic scene understanding, currently incur latencies and resource demands incompatible with safety-critical decision loops. Together, these results establish that cooperative autonomy cannot be advanced through algorithmic sophistication alone, but requires systematic exploration of the joint design space of models, placement, and networks.

Beyond these empirical findings, CADET contributes a methodological foundation for distributed robotics research. By reducing complex multi-node experiments to concise configuration files, the platform transforms previously ad-hoc engineering into reproducible systems experimentation. Its architecture generalizes beyond AVs: drone swarms coordinating via edge nodes, warehouse robots sharing world models through local clusters, and agricultural robots leveraging intermittent satellite connectivity, all face the same entanglement of computation, communication, and coordination. CADET makes these entanglements visible by providing synchronized model-, system-, and task-level instrumentation, enabling causal attribution from design choices to safety outcomes. By releasing CADET as open source, we seek to democratize access to rigorous, systems-aware evaluation of cooperative autonomy, accelerating progress toward robotic systems that are not only intelligent in isolation but robust when distributed across heterogeneous, networked infrastructure. Future work includes extending evaluations to larger multi-vehicle deployments and incorporating additional safety metrics beyond time-to-collision.

% \subsection{Key Insights: Systems Constraints Reshape Algorithmic Choices}
% Across all experiments, CADET reveals a consistent theme: optimal cooperative autonomy strategies emerge from systems-level constraints rather than algorithmic sophistication. V2V's superiority stems from eliminating computation rather than optimizing it. Infrastructure assistance succeeds or fails based on tail latencies beyond algorithmic control. Scalability limits arise from queuing dynamics rather than model accuracy. Foundation models remain viable only when sized for edge deployment, not datacenter scale.
% These discoveries—impossible without CADET's systematic control over distributed execution, network behavior, and resource allocation—demonstrate that advancing cooperative autonomy requires platforms that capture the full complexity of cyber-physical systems. By reducing experimental complexity from thousands of lines of custom code to declarative configurations, CADET democratizes access to systems-aware cooperative autonomy research, enabling the community to explore the vast design space where communication, computation, and coordination intersect.

\section*{Acknowledgment}

This research was funded in part by DEVCOM ARL under the cooperative agreement W911NF1720196, and by the NSF under awards CNS-2211301 and CNS-2325956.

\bibliographystyle{ieeetr}
\bibliography{ieeeconf/sections/ref}

\end{document}